\documentclass[9pt]{article}
\usepackage{spconf}
\usepackage{times}
\usepackage{epsfig}
\usepackage{graphicx}
\usepackage{amsmath}
\usepackage{amssymb}

\usepackage[utf8]{inputenc} %
\usepackage[T1]{fontenc}    %
\usepackage{url}            %
\usepackage{booktabs}       %
\usepackage{amsfonts}       %
\usepackage{bm}             %
\usepackage{nicefrac}       %
\usepackage{microtype}      %
\usepackage{subfig}
\usepackage{multirow}
\usepackage{makecell}
\usepackage{color}
\usepackage[title]{appendix}
\usepackage{fancyhdr}
\usepackage[square,numbers,sort,compress]{natbib}

\usepackage{xcolor}

\newcommand{\func}[1]{\texttt{#1}}

\newcolumntype{L}[1]{>{\raggedright\let\newline\\\arraybackslash\hspace{0pt}}m{#1}}
\newcolumntype{C}[1]{>{\centering\let\newline\\\arraybackslash\hspace{0pt}}m{#1}}
\newcolumntype{R}[1]{>{\raggedleft\let\newline\\\arraybackslash\hspace{0pt}}m{#1}}

\DeclareMathOperator*{\argmin}{arg\,min}

\usepackage[pagebackref=true,breaklinks=true,letterpaper=true,bookmarks=false]{hyperref}

\usepackage{cleveref}

\usepackage{tikz}
\usetikzlibrary{backgrounds, calc, chains, fit, matrix, positioning, shadows, shapes, circuits.ee}
\tikzset{input/.style={}}
\tikzset{output/.style={}}
\tikzset{op/.style={circle, draw, thick, fill=black!10, minimum size=2.5ex, inner sep=0.1ex}}
\tikzset{filter/.style={rectangle, draw, thick, fill=black!10, minimum size=3.5ex, inner sep=1ex}}
\tikzset{other/.style={rounded rectangle, draw, fill=white, minimum size=3.5ex, inner xsep=1ex}}
\tikzset{nn/.style={trapezium, trapezium angle=80, draw, thick, fill=black!10, inner sep=1ex}}
\tikzset{branch/.style={circle, draw, thick, fill=black, minimum size=.5ex, inner sep=0ex}}
\tikzset{tensor/.style={rectangle, draw, thick, fill=white, minimum size=2em, double copy shadow={shadow xshift=.5ex,shadow yshift=-.5ex}}}
\tikzset{image/.style={rectangle, draw, thick, fill=white, minimum size=2em}}
\tikzset{block/.style={rectangle, draw, fill=white, minimum size=2em}}
\tikzset{>=direction ee}

\usepackage{pgfplots}
\pgfplotsset{compat=1.14}
\pgfplotsset{every axis/.append style={enlargelimits={abs=3pt},grid,axis lines=left}}
\pgfplotsset{every axis plot/.append style={thick,mark size=1.5pt,line join=bevel,mark options={solid}}}
\pgfplotsset{label style={font=\small}}
\pgfplotsset{tick label style={font=\footnotesize}}
\pgfplotsset{grid style={color=black!10}}
\pgfplotsset{legend style={draw=none,opacity=.85,font=\footnotesize,cells={anchor=west,opacity=1}}}
\pgfplotsset{every non boxed x axis/.style={xtick align=center,shorten <=-.5\pgflinewidth}}
\pgfplotsset{every non boxed y axis/.style={ytick align=center,shorten <=-.5\pgflinewidth}}
\pgfplotsset{every non boxed z axis/.style={ztick align=center,shorten <=-.5\pgflinewidth}}
\pgfplotsset{/pgf/number format/1000 sep={\,}}

\definecolor{gblue}{HTML}{1f77b4}
\definecolor{ggreen}{HTML}{2ca02c}
\definecolor{bgred}{HTML}{eb6c63}

\title{End-to-end Learning of Compressible Features}
\name{\shortstack{Saurabh Singh$^\dagger$, Sami Abu-El-Haija$^*$, Nick Johnston, Johannes Ballé, Abhinav Shrivastava$^*$,\\ George Toderici}\thanks{$^\dagger$saurabhsingh@google.com. $^*$Work done while at Google.}}
\address{Google Research}
\begin{document}
\maketitle

\begin{abstract}
  Pre-trained convolutional neural networks (CNNs) are powerful off-the-shelf feature generators and have been shown to perform very well on a variety of tasks. Unfortunately, the generated features are high dimensional and expensive to store: potentially hundreds of thousands of floats per example when processing  videos. Traditional entropy based lossless compression methods are of little help as they do not yield desired level of compression, while general purpose lossy compression methods based on energy compaction (e.g. PCA followed by quantization and entropy coding) are sub-optimal, as they are not tuned to task specific objective. We propose a learned method that jointly optimizes for compressibility along with the task objective for learning the features. The plug-in nature of our method makes it straight-forward to integrate with any target objective and trade-off against compressibility. We present results on multiple benchmarks and demonstrate that our method produces features that are an order of magnitude more compressible, while having a regularization effect that leads to a consistent improvement in accuracy.

\end{abstract}

\begin{keywords}
Feature compression, Neural networks
\end{keywords}

\fancyhf{}
\renewcommand{\headrulewidth}{0pt}
\lfoot{\color{gray}\scriptsize Copyright 2020 IEEE. Published in the IEEE 2020 International Conference on Image Processing (ICIP 2020), scheduled for 25-28 October 2020 in Abu Dhabi, United Arab Emirates. Personal use of this material is permitted. However, permission to reprint/republish this material for advertising or promotional purposes or for creating new collective works for resale or redistribution to servers or lists, or to reuse any copyrighted component of this work in other works, must be obtained from the IEEE. Contact: Manager, Copyrights and Permissions / IEEE Service Center / 445 Hoes Lane / P.O. Box 1331 / Piscataway, NJ 08855-1331, USA. Telephone: + Intl. 908-562-3966.}
\thispagestyle{fancy}

\section{Introduction}

Convolutional neural networks (CNNs) have been hugely successful in computer vision and machine learning and have helped push the frontier on a variety of problems~\cite{chen2016deeplab,FasterRCNN,RCNN,papandreou2017towards,SimonyanZ14,AgrawalGM14a,jain201515}. Their success is attributed to their ability of learning a hierarchy of features ranging from very low level image features, such as lines and edges, to high-level semantic concepts, such as objects and parts~\cite{yosinski2015understanding,zeiler2014visualizing}.
As a result, pre-trained CNNs have been shown to be very powerful as off-the-shelf feature generators~\cite{razavian2014cnn}. \citet{razavian2014cnn} demonstrated that a pre-trained network as a feature generator, coupled with a simple classifier such as a SVM or logistic regression, performs surprisingly well and often outperforms hand tuned features on a variety of tasks. This is observed even on tasks that are very different from the original tasks that the CNN was trained on. As a result, CNNs are being widely used as a feature-computing module in larger computer vision and machine learning application pipelines such as video and image analysis. Although features tend to be smaller in size in comparison to the original data they are computed from, their storage can still be prohibitive for large datasets. For example, the YouTube-8M Dataset~\cite{abu2016youtube} requires close to two terabytes of disk space. Compounded by the fact that disk reads are slow, training large pipelines on such datasets becomes slow and expensive. We propose a method to address this issue. Our method jointly optimizes for the original training objective as well as compressibility, yielding features that are as powerful but only require a fraction of the storage cost.

Features also need to be pre-computed and stored for certain types of applications where the target task evolves over time.
A typical example is an indexing system or a content analysis system where the image features may be one of the many signals that the full model relies on. Such a model may change over time by improving how it integrates various signals. It becomes prohibitively expensive to compute features and continuously train such a system end-to-end. While pre-computing features speeds up training, the storage of these features can exceed petabytes for internet scale applications. Our method enables such systems to operate at a fraction of the cost without sacrificing the performance on the target tasks.

CNN features are typically derived by removing the top few layers and using the activations of the remaining topmost layer.
These features tend to be very high dimensional, taking up a significant amount of storage space, especially when computed at a large scale. For example, \cite{abu2016youtube} computes features for 8 million videos and mentions that the original size consisted of hundreds of terabytes, implying that uncompressed features for even a small fraction of YouTube would require hundreds of petabytes. Off-the-shelf lossy compression methods are undesirable for such data as they are content agnostic, resulting in unwanted distortions in the semantic information and a loss of the discriminative power of the original features.

\noindent
\textbf{Contributions:}
We present a method that jointly optimizes for compressibility as well as the target objective used for learning the features. We introduce a penalty that enables a tradeoff between compressibility and informativeness of the features. The plug-in nature of our method makes it easy to integrate with any target objective. We demonstrate that our method produces features that are orders of magnitude more compressible in comparison to traditional methods, while having a regularization effect leading to a consistent improvement in accuracy across benchmarks.

\section{Compressible feature learning}

In a typical supervised classification or regression problem, we are concerned with minimizing a loss function $L$ over a set of parameters $\bm \theta$:
\begin{equation} \label{eq:taskloss}
\bm \theta^\ast = \argmin_{\bm \theta} \sum_{\bm x, \bm y \in \mathcal D} L\bigl(\bm{\hat y}, \bm y \bigr) \qquad \text{ with } \bm{\hat y} = f(\bm x; \bm \theta),
\end{equation}
where $\bm x$ is the input variable (e.g., image pixels, or features), $\bm y$ the target variable (e.g., classification labels, or regression target), $\bm{\hat y}$ is the prediction, $f$ is often an artificial neural network (ANN) with parameters $\bm \theta$ comprising its set of filter weights, and $\mathcal D$ is a set of training data.

Applications using such pre-trained neural networks as feature generators typically remove the top few layers and use the output of remaining topmost layer. We refer to this output as $\bm z$. For the application, $\bm z$ is a set of representative features that can be used in place of $\bm x$. We represent this process of construction of $\bm z$ by splitting $f$ into two parts, $f_z$ and $f_{\hat{y}}$. The prediction $\bm{\hat y}$ is then given by:
\begin{equation}
\bm{\hat y} = f(\bm x; \bm \theta) = f_{\hat{y}}(\bm z; \bm \theta_{\hat{y}}) \text{ with } \bm z = f_z(\bm x; \bm \theta_z) 
\end{equation}
We are interested in learning a model $f$ such that $\bm z$ are compressible while still maintaining the performance of $f$ on the original classification or regression task. Our method achieves this by augmenting the original loss $L$ in \cref{eq:taskloss} with a compression loss $R$ to yield the following optimization problem
\begin{equation} \label{eq:augtaskloss}
\bm \theta^\ast = \argmin_{\bm \theta} \sum_{\bm x, \bm y \in \mathcal D} L\bigl(\bm{\hat y}, \bm y \bigr) + \lambda R(\bm z),
\end{equation}
with $\bm{\hat y}$ and $\bm z$ defined as before and $\lambda$ serving as a trade-off parameter. $R$ encourages the compressibility of $\bm z$ by penalizing an approximation of its entropy as described in the next section. Refer to \cref{app:overview} and \cref{fig:overview} in appendix for additional details.

\subsection{Learning compressible $\bm z$}
General data compression maps each possible data point to a variable length string of symbols (typically bits)~\cite{Sh48}, storing or transmitting them, and inverting the mapping at the receiver side. The optimal number of bits needed to store a discrete-valued data set $\mathcal Z$ is given by the Shannon entropy
\begin{equation}
H = -\sum_{\bm{\hat z} \in \mathcal Z} \log_2 p(\bm{\hat z}),
\end{equation}
where $p$ is a prior probability distribution of the data points, which needs to be available to both the sender and the receiver. The probability distribution is used by \emph{entropy coding} techniques such as arithmetic coding~\cite{RiLa81} or Huffman coding~\cite{van-leeuwen1976} to implement the mapping. The entropy is also referred to as the \emph{bit rate} ($R$) of the compression method.

It is common to speak of an intermediate representation like $\bm z$ as a \emph{bottleneck}. In many cases, for example in the context of autoencoders~\cite{hinton2006}, a bottleneck serves to reduce dimensionality without compromising predictive power, i.e., $\bm z$ is forced to have a smaller number of dimensions than $\bm x$. Since the number of dimensions is a hyperparameter (an architectural choice), no changes to the loss function are necessary, and the model is simply trained to minimize the loss under the given constraint. However, dimensionality reduction is only a crude approximation to data compression.

In data compression, the compressibility of a dataset can be greatly increased by allowing errors in the compression process -- referred to as \emph{lossy compression}. A lossy compression method allows the bit rate $R$ to be traded off against the distortion $D$ introduced in the data. This rate-distortion trade-off is represented as the following optimization problem.
\begin{equation} \label{eq:ratedistortion}
\bm{\hat z}^\ast = \argmin_{\bm{\hat z}} \sum_{\bm x \in \mathcal D} D(\bm{\hat z}, \bm x) + \lambda R(\bm{\hat z}),
\end{equation}
where $\bm{\hat z}$ is a discrete lossy representation of $\bm x$. Note the similarity between \cref{eq:augtaskloss} and \cref{eq:ratedistortion}. The key difference is that while \cref{eq:ratedistortion} measures the distortion in data directly, \cref{eq:augtaskloss} measures the ``distortion'' using the target variable $\bm y$ and ignores the input $\bm x$. Our key observation is that the supervised losses, such as  $L(\bm{\hat y}, \bm y)$ used for classification, can take the place of distortion in \cref{eq:ratedistortion}. We therefore re-cast our original optimization problem in \cref{eq:augtaskloss} as the following rate-distortion optimization problem
\begin{equation} \label{eq:entropyaugmented}
\bm \theta^\ast, \bm \phi^\ast = \argmin_{\bm \theta, \bm \phi} \sum_{\bm x, \bm y \in \mathcal D} \underbrace{L(\bm{\hat y}, \bm y)}_{\text{distortion} D} + \lambda \cdot \underbrace{-\log_2 p\bigl(\bm{\hat z}; \bm \phi\bigr)}_{\text{bit rate} R},
\end{equation}
where $\bm{\hat z} = \lfloor f_z(\bm x; \bm \theta_z) \rceil$, $\bm{\hat y} = f_{\hat{y}}(\bm{\hat z}; \bm \theta_{\hat{y}})$, and $p$ is a probability model over $\bm{\hat z}$ with parameters $\bm \phi$, which are trained jointly with $\bm \theta \equiv \{\bm \theta_z, \bm \theta_{\hat{y}}\}$. $\left\lfloor{\cdot}\right\rceil$ here indicates that we round the output of $f_z$ to the nearest integers. This quantization is necessary, as compression can take place only in a discrete space with a countable number of possible states. Note that this is how a trade-off between compression performance and prediction performance is achieved. By reducing the number of possible states in $\bm{\hat z}$, for example by scaling down the outputs of $f_z$, the bit rate $R$ can be reduced at the expense of prediction performance. On the other hand, the prediction performance can be improved by increasing the number of possible states, at the expense of compressibility. The hyperparameter $\lambda$ controls trade-off. We call this type of bottleneck an \emph{entropy bottleneck}.

\subsection{Optimizing with discrete $\bm{\hat z}$}
It is not feasible to minimize the objective in \cref{eq:entropyaugmented} directly with descent methods, as the quantization leads to gradients that are zero almost everywhere. Instead, we closely follow the approach introduced in \cite{BaLaSi16a} with one key difference. \citet{BaLaSi16a} substitute the quantization with additive uniform noise during training, while we do so only for modeling the rate. For distortion, we discretize (by rounding) and substitute the gradients by identity (straight-through). Further, rather than using a piecewise linear density model as in \cite{BaLaSi16a}, we use a more refined density model which is described in \cite{balle2018variational}.

For all the experiments in this paper, a separate model was used for each vector element $\hat z_i$ in $\bm{\hat z}$, yielding a fully factorized probability model 
$p(\bm{\hat z}) = \prod_i p(\hat z_i)$.
For bottlenecks with a spatial configuration, all the spatial elements within the same channel share the distribution. 

\section{Experiments}

\begin{figure*}[t]
\centering
\subfloat[CIFAR-10]{\label{fig:cifar10baseline}
\includegraphics[width=0.48\textwidth]{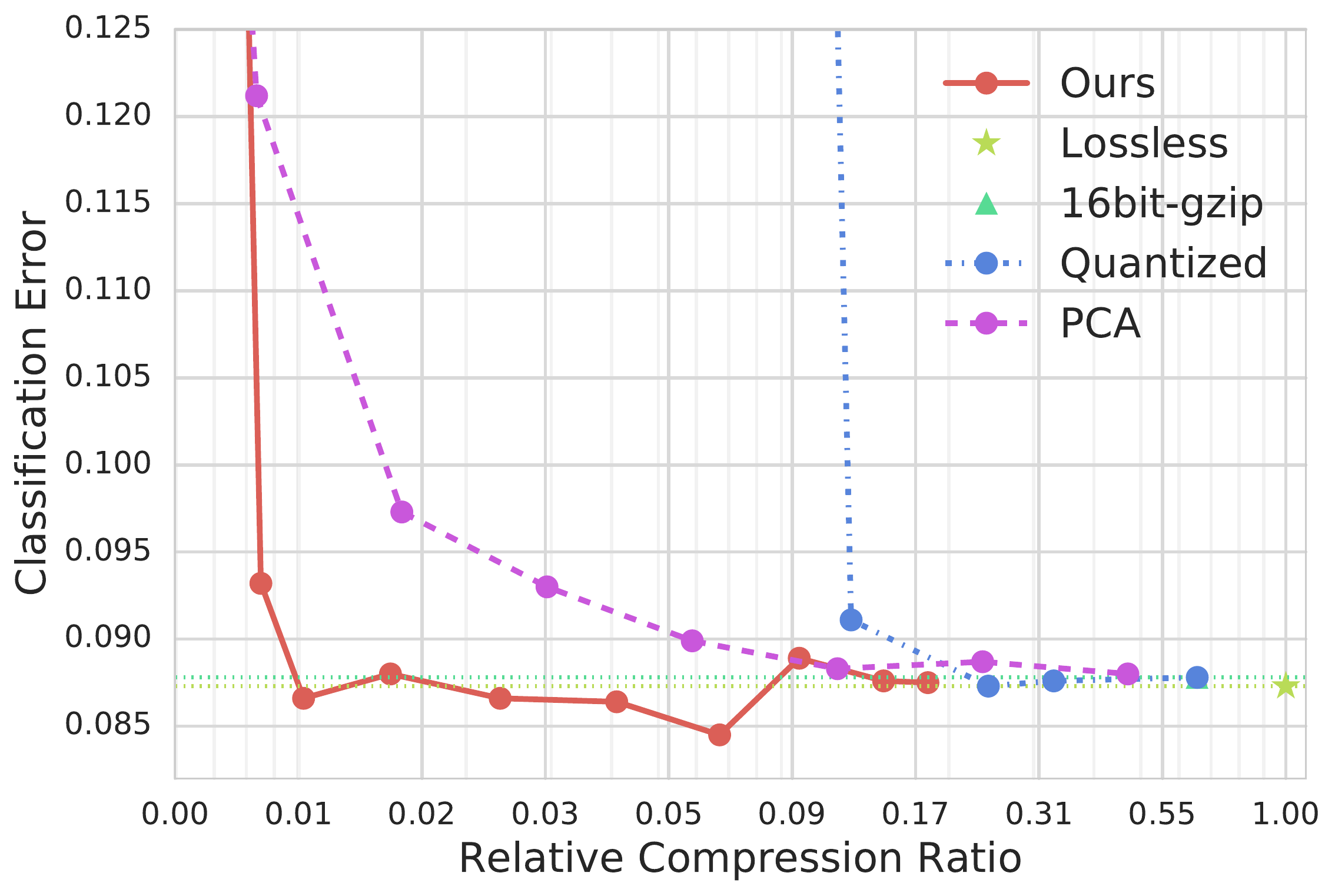}
}
\subfloat[CIFAR-100]{\label{fig:cifar100baseline}
\includegraphics[width=0.48\textwidth]{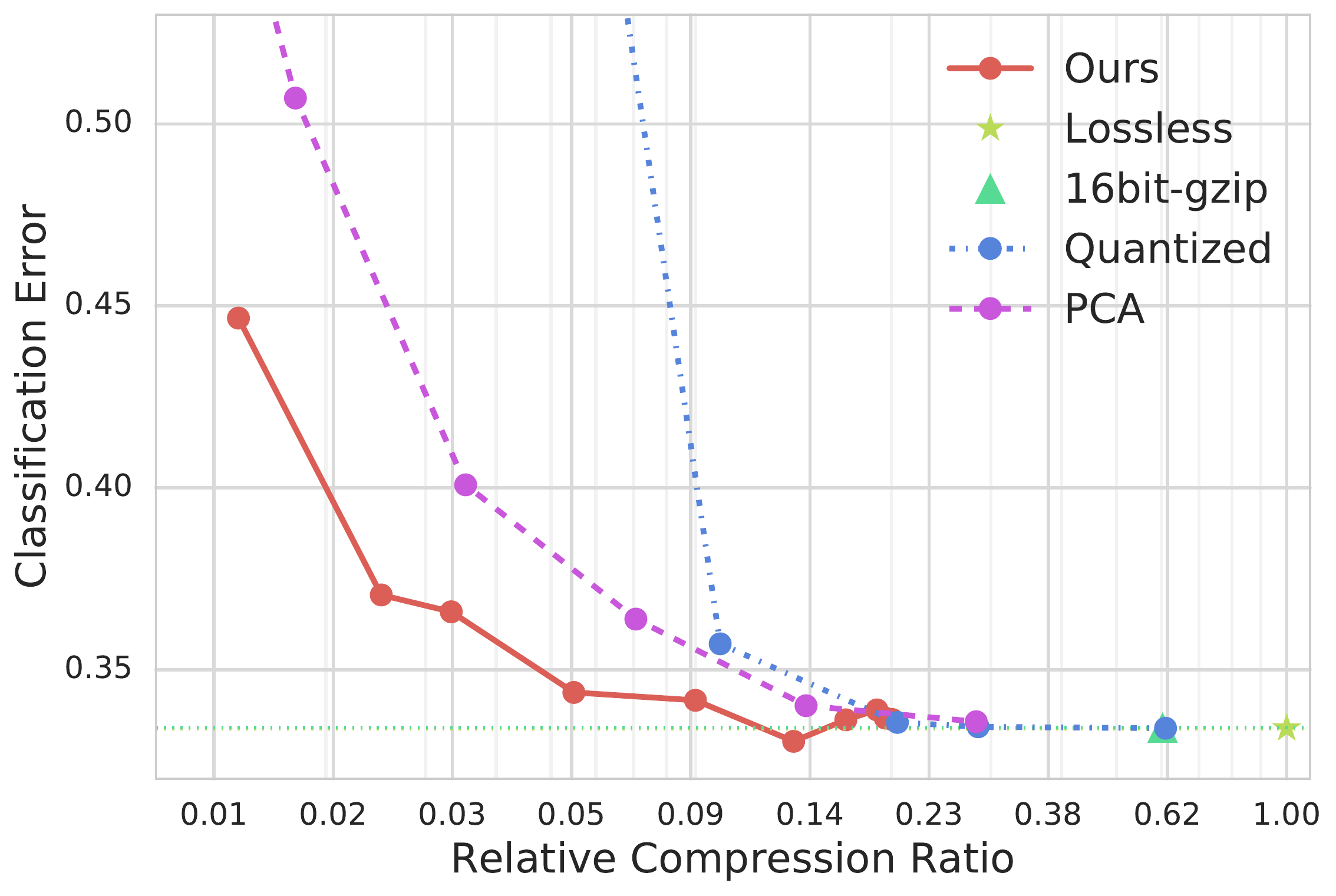}
}
\hfill
\subfloat[Imagenet]{\label{fig:ImageNet_baselines}
\includegraphics[width=0.48\textwidth]{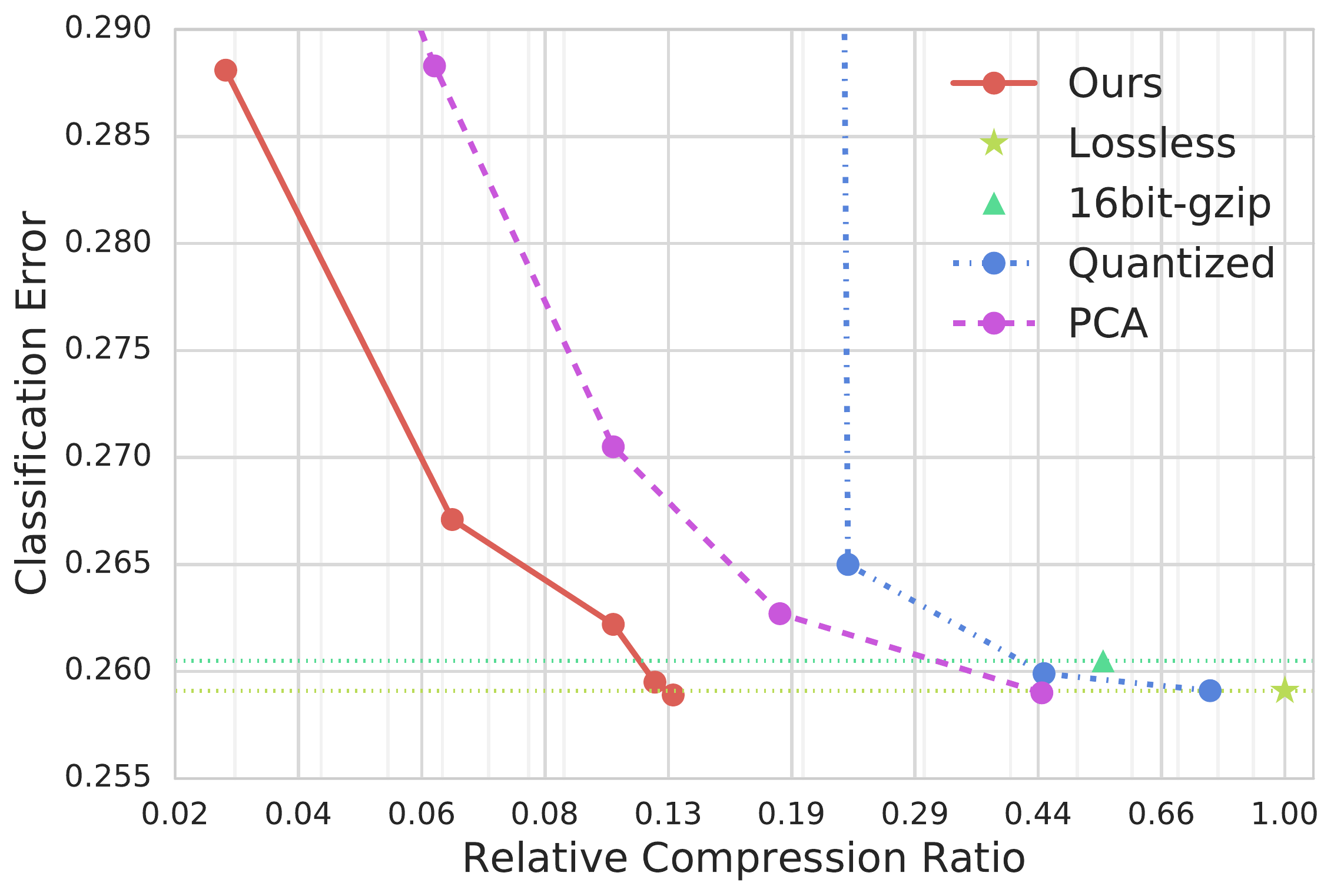}
}
\subfloat[YouTube-8M]{\label{fig:yt8m_baselines}
\includegraphics[width=0.48\textwidth]{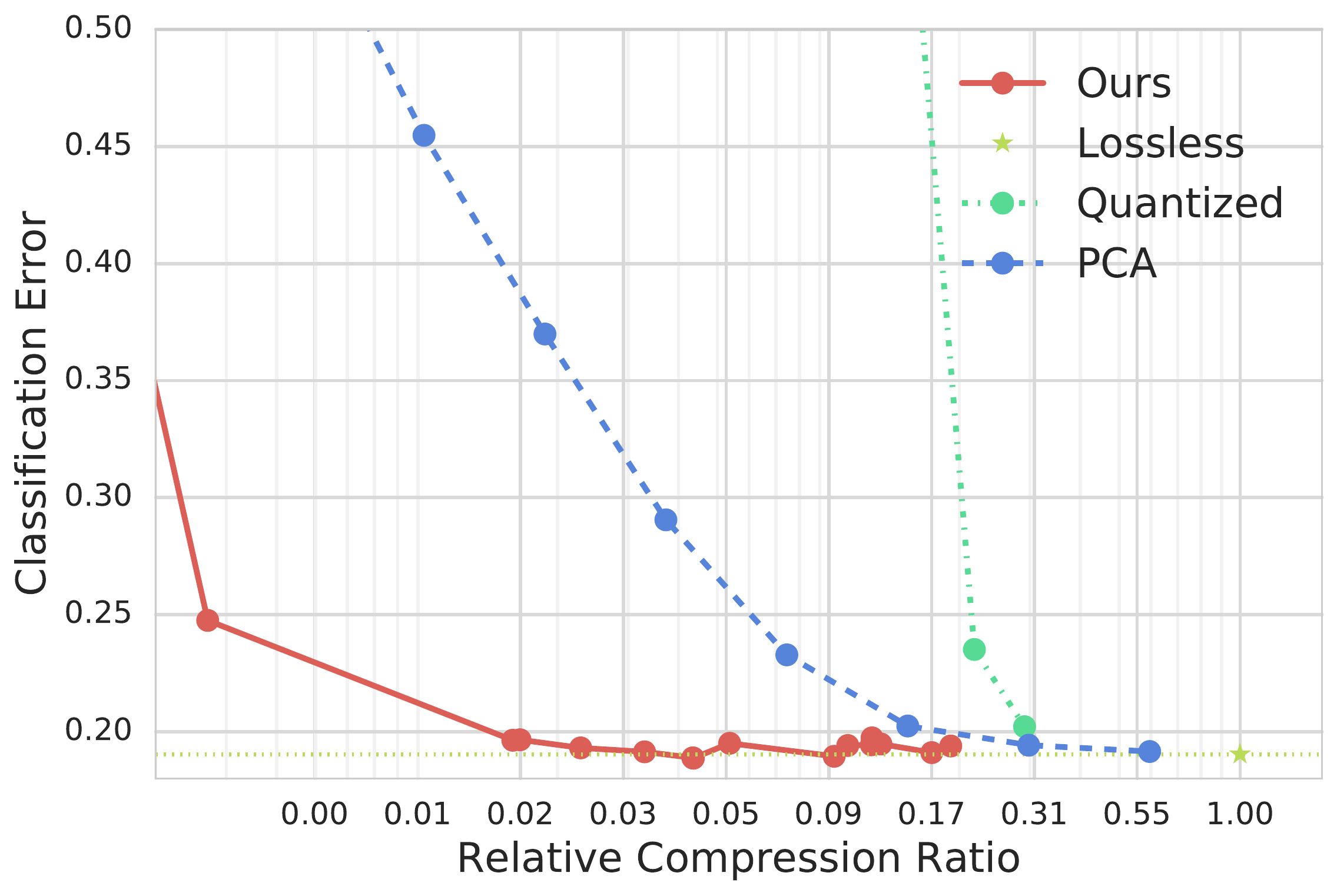}
}
\caption{We visualize the
classification error of the decompressed features as a function of the relative compression ratio with respect
to the lossless compression on CIFAR-10~(\ref{fig:cifar10baseline}), CIFAR-100~(\ref{fig:cifar100baseline}), ImageNet~(\ref{fig:ImageNet_baselines}) and YouTube-8M~(\ref{fig:yt8m_baselines}).
On CIFAR-10 our method produces representations that preserve the accuracy at 1\% the size of the losslessly compressed size, while on CIFAR-100 at 10\% the losslessly compressed size. On both ImageNet and YouTube-8M, our method preserves the accuracy while reducing the storage cost to $\approx13\%$ and 10\% of the losslessly compressed file size respectively.
}
\label{fig:cifar_baselines}
\end{figure*}

We evaluate our method using classification models, as they are the most common off-the-shelf feature generation method. Unless otherwise stated, in all the following experiments, we follow the standard practice of considering the activations of the penultimate layer immediately after the non-linearity as the feature layer. We treat it as the bottleneck $\bm{\hat z}$ and apply the rate penalty over it. We train several models by varying the trade-off parameter $\lambda$ and present our results in the form of a error vs. relative compression graph. Relative compression is measured as a fraction of the compressed size achieved by lossless compression baseline \func{zlib} described below. For all the methods, the representation for each image is computed in \func{float32} precision and compressed independently. Additional details are provided in \cref{app:trainingdetails} of the appendix.

\subsection{Baseline compression methods}
We compare our method with the following standard baselines.

\noindent
\textbf{Lossless compression:} The features are compressed using the gzip compatible \func{zlib} compression library in Python. The representation is first converted to a byte array and then compressed using \func{zlib} at the highest compression level of 9.

\noindent
\textbf{16bit-gzip compression:} The features are first cast to a 16 bit floating point representation and then losslessly compressed using \func{zlib} as described above.

\noindent
\textbf{Quantized:} The features are scaled to a unit range followed by quantization to equal length intervals. We report performance as a function of the number of quantization bins in the set $\{2^{16}, 2^8, 2^4, 2^2\}$. These quantized values are losslessly compressed using gzip as above. If fewer than 256 quantization bins are used, the data is natively stored as a byte, not packed, before gzip is used.

\noindent
\textbf{PCA:} We compute principal components from the full covariance matrix of the features computed over the entire training set. We report the performance as a function of the number of components used from the set  $\{1, 2, 4, 8, 16, 32, 64\}$. We exclude the cost of PCA basis from the compression cost.

\subsection{Evaluation on CIFAR-10/100}
\noindent
\textbf{Setup:}
CIFAR-10 and CIFAR-100 image classification datasets contain 10 and 100 classes respectively. Both
contain 50000 training and 10000 testing images. We use a 20 layer ResnetV2~\cite{resnetsv2} model and train using SGD with ADAM~\cite{kingma2014adam} for 128k iterations with a batchsize of 128.

\noindent
\textbf{Results:}
Figure~\ref{fig:cifar_baselines} shows that our method consistently produces features that are an order of magnitude more compressible than when losslessly compressed, while maintaining the discriminative power of the learned features. We visualize the classification error of the decompressed features as a function of the relative compression ratio with respect to the lossless compression. On CIFAR-10 our method is able to produce features that are 1\% the size of the lossless compression while matching the accuracy. This is likely due to the fact that there are only $10$ classes which would ideally only require $\log_2{10}$ bits. For CIFAR-100, we observe that our method produces features that can be compressed to 10\% the size of lossless compression while retaining the same accuracy. Here we see an order of magnitude reduction in achieved compression in comparison to CIFAR-10 with an order of magnitude increase in number of categories (from 10 to 100). On both the datasets 16bit-gzip consistently retains performance indicating that 16bit precision is accurate enough for these features. Quantization quickly loses performance as the number of quantization bins is decreased. PCA performs better than other baselines. However, its performance quickly degrades as fewer components are used. The results summarized in Table \ref{tab:generalization} also show that the best performing rate points on the validation set also exhibit a higher training error than the baseline. This is an indication of the regularization effect of our method.

\subsection{Evaluation on ImageNet}
\noindent
\textbf{Setup:}
We train on $\approx 1.2M$ training images and report results on the 50000 validation images in the Imagenet classification dataset~\cite{ImageNet}. We use a 50 layer ResnetV2~\cite{resnetsv2} model as our base model. All networks are trained using SGD with ADAM~\cite{kingma2014adam} for 300k steps using a batchsize of 256.

\noindent
\textbf{Results:} Our method produces highly compressible representations in comparison to the other baseline methods and is able to preserve the accuracy while reducing the storage cost to $\approx 12.2\%$ of the losslessly compressed file size  (25.95\% Ours vs. 25.91\% Lossless). Note that lossless storage at 16bit precision results in a 0.14\% increase in error. Similar to CIFAR-10/100 datasets, we observe a regularization effect. As evident in Table~\ref{tab:generalization}, despite a higher error on the training set in comparison to the baseline, validation performance improves.

\subsection{Evaluation on Youtube 8 Million Dataset}

\noindent
\textbf{Setup:}
YouTube-8M \cite{abu2016youtube} is one of the largest publicly available video classification dataset. We use the second version (v2), which contains 6.1 million videos and 3862 classes. We first aggregate the video sequence features into a fixed-size vector using mean pooling. We use a three fully-connected layer model, with ReLU activation and batch normalization on the hidden layers. We apply the compression on the last hidden activations, just before the output layer. Models are trained using TensorFlow's Adam Optimizer \cite{kingma2014adam} for 300,000 steps using a batchsize of 100.

\noindent
\textbf{Results:}
Figure \ref{fig:yt8m_baselines} and Table~\ref{tab:generalization} report the accuracy on the validation partition of YouTube-8M \cite{abu2016youtube}. Similar to other benchmarks, our method can drastically lower the storage requirements while preserving and even improving the accuracy, providing further evidence for the regularization effect. Refer to \cref{app:youtube8m} in appendix for additional results and discussion.

\subsection{Additional Discussion}
Please refer to \cref{app:discussion} in appendix for additional details and discussion.

\begin{table*}[t]
  \caption{We compare the total compressed size of the evaluation datasets along with the final training and validation errors. For each dataset we select the lowest rate model with error lower than baseline. For YouTube-8M the reported size is of video level features. The gap between train and validation errors is consistently smaller for our model,
  indicating that the entropy penalty has a regularization effect. At the same time, our model significantly reduces the total size.}
  \label{tab:generalization}
  \centering
  \begin{tabular}{@{}L{2cm} r r r r r r r@{}}
    \toprule
    & 
    \multicolumn{2}{c}{Training Error} & 
    \multicolumn{2}{c}{Validation Error} & 
    \multicolumn{3}{c}{Validation Set Size}\\
    \cmidrule(l{0.3em}r{0.5em}){2-3}
    \cmidrule(l{0.2em}r{0.6em}){4-5}
    \cmidrule(l{0.5em}r{0.2em}){6-8}
    & \multicolumn{1}{c}{Lossless} & 
    \multicolumn{1}{c}{Ours} &
    \multicolumn{1}{c}{Lossless} & 
    \multicolumn{1}{c}{Ours} & 
    \multicolumn{1}{c}{Lossless} & 
    \multicolumn{1}{c}{Ours} & 
    \multicolumn{1}{c}{Raw}  \\
    \midrule
    ImageNet  & 17.04 & 17.35 & 25.91 & 25.89 & 6.95GB & 0.85GB & 38.15GB \\
    CIFAR-10  & 0.14 & 0.29 & 8.73 & 8.45 & 41.53MB & 2.78MB & 156.25MB \\
    CIFAR-100 & 0.25 & 1.54 & 33.39 & 33.03 & 69.14MB & 9.28MB & 156.25MB \\
    YouTube-8M & 19.56  & 19.75 & 19.76 & 19.49 & 5.30GB & 0.27GB & 17.80GB \\
    \bottomrule
  \end{tabular}
\end{table*}

\section{Related Work}
\noindent
\textbf{Representations from off-the-shelf compression algorithms:}
Compressed representations have been directly used for training machine learning algorithms as they have low memory and computational requirements and enable efficient real-time processing while avoiding decoding overhead.
\citet{aghagolzadeh2015hyperspectral} used linear SVM classifier for pixel classification on compressive hyperspectral data. Hahn et al.~\cite{hahn2014adaptive} performed hyperspectral pixel classification on the compressive domain using an adaptive probabilistic approach. Fu et al.~\cite{fu2016using} fed DCT compressed image data into the network to speed up
machine learning algorithms applied on the images.
Biswas et al.~\cite{biswas2013h} proposed an approach to classify H.264 compressed videos. Chadha et al.~\cite{chadha2017video} used 3D CNN architecture for video classification that directly utilized compressed video bitstreams. Yeo et al.~\cite{yeo2008high} designed a system for performing action recognition on videos compressed with MPEG. Kantorov et al.~\cite{kantorov2014efficient} proposed a method for extracting and encoding local video descriptors for action recognition on MPEG compressed video representation.
Our work differs in that we jointly optimize for a compressible representation along with the target task.

\noindent
\textbf{Joint optimization for compression:}
\citet{torfason2018towards} propose to extend an auto-encoding compression network by adding an additional inference branch over the bottleneck for auxiliary tasks. Our method does not use any auto-encoding penalty and directly optimizes for the entropy along with the task specific objective.

\noindent
\textbf{Dimensionality reduction methods:} 
While not performing information-theoretic compression, there are many lossy dimensionality reduction methods which can be applied to reduce the space needed for precomputed CNN features. For example, PCA, LDA, ICA, Product Quantization~\cite{jegou2011product} etc.
However, none of these methods takes into account the task specific loss. Instead they all rely
on surrogate losses (e.g. $L_2$).

\noindent
\textbf{Similarity preserving hashing:}
Hashing based methods have been used to produce a neighborhood preserving compact binary embedding of the data~\cite{liu2016hash, lai2015hash, xia2014hash, zhao2015hash}. 
Such methods are similar to compression in that a binary representation smaller than the data itself is found. However, the storage size is a hyperparameter that is not directly optimized and representations are typically of identical length with the goal of minimizing lookup speed as opposed to storage. Refer to \cref{app:simdiff} in appendix for further details.

\noindent \\
\textbf{Variational information bottleneck:}
Our approach can be viewed as a particular instantiation of the more general information bottleneck framework~\cite{tishby2000information, alemi2016vib}. While these works discuss mutual information as the parameterization independent measure of informativeness, we note that a task dependent measure can typically be used and may be better suited if target tasks are known ahead of time. As most classification models are typically trained to optimize cross entropy, we use the same in this paper as a measure of informativeness.

\section{Conclusion}
We presented an end-to-end trained method to learn compressible features while training for a task dependent objective. By evaluating on four different benchmarks we demonstrated that our method achieves high compression rates compared to classical methods, while having a regularization effect leading to a consistent improvement in accuracy across benchmarks.

{\small
\bibliographystyle{unsrtnat}
\bibliography{paper.bib}
}

\vfill
\clearpage

\twocolumn[  
    \begin{@twocolumnfalse}
        \begin{center}
             {\Large\bfseries{\scshape{Appendix \& Supplemental Material}}}
        \end{center}
        \vspace{8mm}
    \end{@twocolumnfalse}
]

\appendixtitleon
\begin{appendices}

\section{Additional Method Details}
\label{app:overview}

\begin{figure}[t]
\resizebox{\linewidth}{!}{
  \begin{tikzpicture}[x=1em,y=1em]
    \node [image] (x) {$\bm{x}$};
    \node [image] (yhat) at ($(x)+(0,-12)$) {$\bm{\hat{y}}$};
    
    \node[nn, shape border rotate=270, minimum size=2em] (f_z) at  ($(x)+(5,0)$) {$f_z(\bm{x}; \bm \theta_z)$};
    \node[nn, shape border rotate=90, minimum size=2em] (f_yhat) at ($(yhat)+(5,0)$) {$f_{\hat{y}}(\bm z; \bm \theta_{\hat{y}})$};
    
    \node [image] (z) at ($(f_z)+(5, 0)$) {$\bm{z}$};
    \node [image] (z_hat) at ($(f_yhat)+(5, 0)$) {$\bm{\hat{z}}$};
    
    \node[op] (noise_op) at ($(z)+(0,-3)$) {$+$};
    \node [input] (u) at ($(f_z)+(0,-3)$) {$u \sim \mathcal{U}[\cdot]$};
    
    \node[op](round_op) at ($(z)+(3,0)$) {\footnotesize $\lfloor \cdot \rceil$};
    \node[filter](ec) at ($(round_op)+(7, -3)$){EC};
    \node[block](bitstream) at ($(ec)+(0, -3)$){\tiny $01001\cdots$};
    \node[filter](ed) at ($(bitstream)+(0, -3)$){ED};
    \node[filter](em) at ($(z_hat)+(0, 6)$){$p(\bm{\hat z}; \bm \phi)$};
    \node[filter, align=center](probtab) at ($(z_hat)+(6, 6)$){Prob.\\Table};
    
    \node [output] (lyy) at ($(yhat)+(0, 3)$) {$L(\bm{\hat y}, \bm y)$};
    \node [output] (rate) at ($(yhat)+(0, 6)$) {$R(\bm{\hat z})$};
    
    \coordinate (hidden) at ($(z_hat)+(3.1,3)$);
    
    \draw[->] (x) -- (f_z) -- (z) -- (round_op);
    \draw[->] (z_hat) -- (f_yhat) -- (yhat);
    \draw[->, ggreen] (u) -- (noise_op);
    \draw[->, bgred] (round_op) -| (ec) -- (bitstream);
    \draw[->, ggreen] (round_op) -- ($(z_hat)+(3,3)$) -- ($(z_hat)+(0,3)$) -- (z_hat.north);
    \draw[->, bgred] (bitstream) -- (ed) |- (z_hat);
    \draw[->, ggreen] (z) -- (noise_op) -- (em) -- (rate);
    \draw[->, thick, densely dotted] (em) -- ($(em.south)+(0,-1.25)$) -| (probtab.260);
    \draw[->, ggreen] (yhat) -- (lyy);
    \draw[->, thick, densely dotted, bgred] (probtab.80) |- (ec);
    \draw[->, thick, densely dotted, bgred] (probtab.280) |- (ed);
    
    \begin{pgfonlayer}{background}
      \node[fill=ggreen, opacity=.2, rounded corners=2ex, fit=(rate)(lyy)(em)(noise_op)(hidden), inner xsep=1.0ex, inner ysep=1.0ex] {};
      \node[fill=bgred, opacity=.2, rounded corners=2ex, fit=(ec)(bitstream)(ed)(probtab), inner xsep=1.0ex, inner ysep=1.0ex] {};
    \end{pgfonlayer}
  \end{tikzpicture}\hfill
  }
  \caption{Overview of our method. During training (green box and arrows), uniform noise $u$ is added to $\bm z$ to simulate quantization while allowing gradient based optimization. An entropy model $p(\bm{\hat z}; \bm \phi)$ is used to model the distribution of $\bm{\hat z}$ and impose compression loss $R(\bm{\hat z})$. During testing (red box and arrows), $\bm{z}$ are quantized using rounding to yield $\bm{\hat z}$ and entropy coding (EC) is then used for lossless compression yielding a variable length bit string for storage and transmission. This bit string can be decoded using entropy decoding (ED) to yield $\bm{\hat z}$ which can then be further processed. Both EC and ED use probability tables produced from the entropy model $p(\bm{\hat z}; \bm \phi)$ after the training is complete. Note that $\bm{\hat z}$ represents noise added $\bm{z}$ as well as quantized $\bm{z}$ depending on the context.}
  \label{fig:overview}
\end{figure}
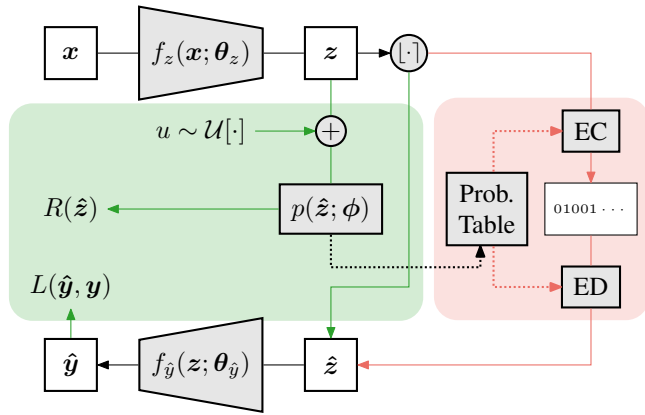

In \cref{fig:overview} we show the model during training using green box and green arrows, while red box and red arrows show the model at test time. Common components are shown outside the colored boxes and use black arrows. Note that, during training uniform noise is added to simulate quantization while during testing rounding is used for quantization. Since entropy coding and decoding are lossless operations, they are not used during training. Also note that the learned entropy model $p(\bm{\hat z}; \bm \phi)$ during training is used to produce probability tables which are used by entropy coding and decoding during testing. These dependencies are shown as dotted arrows in the figure. Once the model is trained and probability tables are produced, the entropy model $p(\bm{\hat z}; \bm \phi)$ is not required and can be discarded.

\section{Additional Discussion}
\label{app:discussion}

\subsection{Importance of Adam optimizer:}
The weight $\lambda$ for the entropy penalty in \cref{eq:entropyaugmented} can affect the magnitude of the gradient updates for the parameters $\phi$ of the probability model. A smaller value of $\lambda$ can reduce the effective learning rate of $\phi$ causing the model to learn slower. This may result in a disconnect between the observed distribution and the model. Adam optimizer~\cite{kingma2014adam} computes updates normalized by the square root of a running average of the squared gradients. This has the desirable property that a constant scaling of loss does not affect the magnitude of updates. Therefore, for the combined loss in \cref{eq:entropyaugmented}, the $\lambda$ only affects the relative weight of the gradient due to the entropy penalty, without changing the effective learning rate of $\phi$.

\subsection{Regularization effect:}
In addition to pure lossy compression, we show that our method has an
interesting side effect: it acts as an activation regularizer, allowing higher classification
results on the validation than the original network,
while exhibiting higher training error (Table~\ref{tab:generalization}). Interestingly the regularization effect's sweet spot may provide some insight in the
complexity of the problem to be solved. Unlike normal
regularization methods, our approach makes a tradeoff between the information
passed between the encoder network and the classifier, therefore we can explicitly
measure how much information is required for a particular classification task.
We observed that CIFAR-100 requires less compression to achieve the best 
result, whereas CIFAR-10 requires about half as much information in order to obtain the best result, which signals that perhaps the network designed to solve both problems
is perhaps a bit larger than it should be in the case of CIFAR-10. 

\subsection{Note on deep compression without decoding~\cite{torfason2018towards}:}
While there are significant differences in the model and setup of \citet{torfason2018towards}, we can qualitatively compare the performance on the ImageNet classification task in terms of the relative increase in error rate versus the baselines at roughly 0.3 bits per pixel (bpp). We observe a \textit{relative increase} of 10\% in error at 0.388 bpp  (corresponding to the highest compression rate in Fig. \ref{fig:ImageNet_baselines}), while in \cite{torfason2018towards}(Table 2), the relative increase reported at 0.330 bpp is 20.3\%, indicating that our model is able to better preserve the informativeness of the features.

\subsection{Differences from similarity preserving hashing}
\label{app:simdiff}

We enumerate the key differences below:
\begin{itemize}
\item In hashing, the binary representations are required to preserve neighborhood to enable direct retrieval based on the hash value. In our method the compressed bits are output of arithmetic coding with no such constraints. 
\item The storage size of the representation is fixed (a hyperparameter) in binary hashing and not directly optimized, while in compression it is directly incorporated in loss (as a rate term, \cref{eq:entropyaugmented}) to tradeoff with accuracy.
\item In hashing, values of identical length are typically produced with lookup speed benefits while in compression storage size is the primary concern and entropy coding is used to produce variable length bit representations.
\end{itemize}

\section{Training Details}
\label{app:trainingdetails}

\subsection{CIFAR-10/100}
We used a cosine decay learning rate schedule~\cite{loshchilov2016sgdr} with an initial learning rate of 0.005 (selected as best among \{0.1, 0.05, 0.025, 0.01, 0.005, 0.0025, 0.001\}). We use the standard data augmentation of left-right flips and zero padding all sides by 4 pixels followed by a $32\times32$ crop. We use a weight decay of 0.0001 and train our model on a single GPU using a batch size of 128. 

\subsection{Imagenet}
We use cosine decay learning rate schedule~\cite{loshchilov2016sgdr} with an initial learning rate of 0.001. We use the standard data augmentation as used in~\cite{googlenet} and train on crops of $299\times299$.  We use a weight decay of 0.0001 and train each model on 8 GPUs with a batch size of 32 per GPU resulting in a combined batch size of 256 and synchronous updates. We report top-1 classification error computed on a $299\times299$ center crop.

\subsection{Youtube 8 Million Dataset}
We use weight decay of $10^{-6}$ and train each model on one CPU, with a batch size of $100$, minimizing cross-entropy loss, using TensorFlow's Adam Optimizer \cite{kingma2014adam} for 300,000 steps. We sweep the initial learning rate from choices $\{0.04, 0.02, 0.002, 0.001, 0.0005\}$, and we multiply the learning rate by $0.94$ every $1000$ steps. For each model architecture, we use the best learning rate according to a held-out validation set.

\begin{figure}[t]
\centering
\subfloat[2-layer model: input $\rightarrow \mathbb{R}^{4000}\rightarrow \mathbb{R}^{3862}$]{\label{fig:yt8m_4000}
\includegraphics[width=0.48\textwidth]{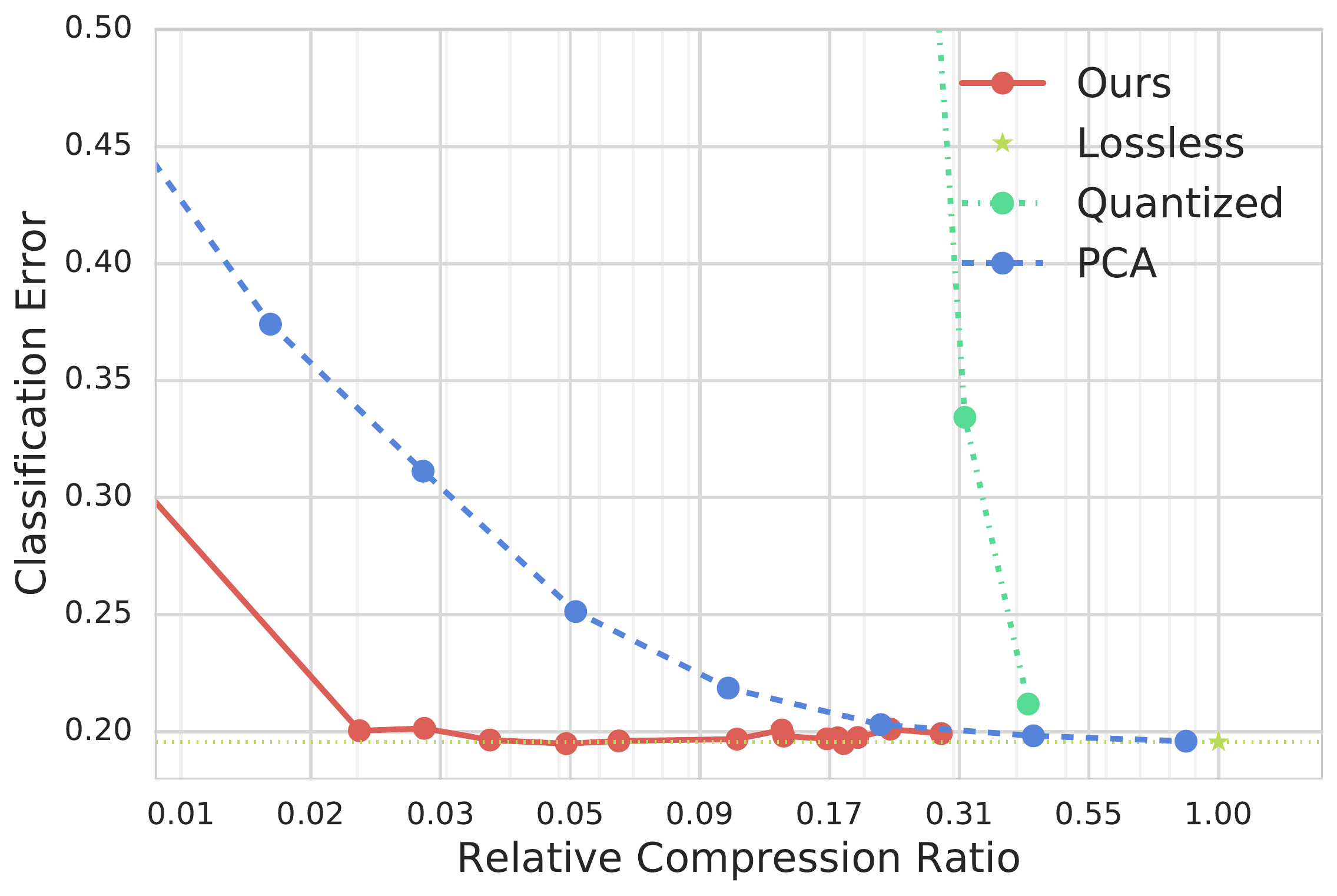}
}
\hfill
\subfloat[3-layer model: input $\rightarrow \mathbb{R}^{6000}\rightarrow \mathbb{R}^{2000}\rightarrow \mathbb{R}^{3862}$]{\label{fig:yt8m_6000x2000}
\includegraphics[width=0.48\textwidth]{figs/yt8m_baseline_6000x2000.pdf}
}
\caption{Evaluation on YouTube-8M. We evaluate our method for two different choices of architectures: (a) 2-layer network and (b) 3-layer network. We observe 5\% to 10\% reduction in the storage cost in comparison to the losslessly compressed file size while preserving accuracy.}.
\label{appfig:youtube8m}
\end{figure}

\section{Additional results on Youtube 8 Million Dataset}
\label{app:youtube8m}
\Cref{fig:yt8m_baselines} in main text presented results for a three layer model. For comparison, we also present the results for a two layer model in \cref{fig:yt8m_4000}. \Cref{fig:yt8m_baselines} is also reproduced as \cref{fig:yt8m_6000x2000} for ease of comparison. As before, we observe that a drastic reduction in the storage requirements while preserving accuracy. However, the three layer model preserves accuracy up to a higher compression ratio than the three layer model. Note that, the accuracy metrics are measured on the ``validation'' partition of YouTube-8M \cite{abu2016youtube}.

Our method can potentially have a large impact on the online video systems. By extrapolation, assume a hypothetical online video system with 1 billion videos, that wishes to annotate the videos using the audio-visual content to support a Video Search application. Storing the mean-pooled audio-visual features (identical to ones available in YouTube-8M, say) would require a storage of 4.6 Terabytes in the raw storage form. However, if the annotation system was trained with our method, then the storage requirements would drop to less than 150 Gigabytes, and yet possibly improve the generalization performance for annotating new videos. 

\end{appendices}
\end{document}